\documentclass[12pt,authoryear]{elsarticle}

\usepackage{amssymb}
\usepackage{amsmath}

\usepackage{url}
\usepackage{booktabs}
\usepackage{xcolor}

\begin{document}

\begin{frontmatter}



 \title{ISSR: Iterative Selection with Self-Review           for Vocabulary Test Distractor Generation}


\author{Yu-Cheng Liu} 
\ead{liu2022113.cs11@nycu.edu.tw}
\affiliation{organization={Department of Computer Science, National Yang Ming Chiao Tung University},
            city={Hsinchu},
            country={Taiwan},}

\author{An-Zi Yen} 
\ead{azyen@nycu.edu.tw}

\begin{abstract}
Vocabulary acquisition is essential to second language learning, as it underpins all core language skills. 
Accurate vocabulary assessment is particularly important in standardized exams, where test items evaluate learners' comprehension and contextual use of words. 
Previous research has explored methods for generating distractors to aid in the design of English vocabulary tests. However, current approaches often rely on lexical databases or predefined rules, and frequently produce distractors that risk invalidating the question by introducing multiple correct options. 
In this study, we focus on English vocabulary questions from Taiwan’s university entrance exams. 
We analyze student response distributions to gain insights into the characteristics of these test items and provide a reference for future research. 
Additionally, we identify key limitations in how large language models (LLMs) support teachers in generating distractors for vocabulary test design. 
To address these challenges, we propose the iterative selection with self-review (ISSR) framework, which makes use of a novel LLM-based self-review mechanism to ensure that the distractors remain valid while offering diverse options. 
Experimental results show that ISSR achieves promising performance in generating plausible distractors, and the self-review mechanism effectively filters out distractors that could invalidate the question.
\end{abstract}




\end{frontmatter}




\section{Introduction}\label{sec:intro}

In second language learning, vocabulary acquisition plays a critical role as it serves as the foundation for all language skills, including listening, speaking, reading, and writing. 
Expanding one's vocabulary is a key factor in enhancing overall language comprehension~\citep{yorio1971some}.
A strong vocabulary enhances learners' ability to comprehend spoken and written texts, as a larger vocabulary enables more effective understanding. 
Additionally, understanding vocabulary is essential for mastering grammar, as knowing the meanings and functions of words aids in grasping sentence structures and applying grammatical rules. 
Therefore, vocabulary acquisition is fundamental to the overall development of language proficiency in second language learners.

Numerous studies have explored various aspects of vocabulary tests in language learning~\citep{morea2024diverse,ha2022issues,tseng2016measuring,abdullah2013validation}.
Given the significance of vocabulary acquisition, it is crucial to develop effective ways of assessing learners' vocabulary knowledge. 
One common method is the use of vocabulary test items that require learners to select the correct word to fill a gap from a set of multiple choices. 
This type of assessment not only measures learners' ability to recognize the meaning of individual words but also evaluates their understanding of how words function in specific contexts.  
Vocabulary questions have become an integral part of standardized English exams, especially in Asia, where they play a significant role in evaluating language proficiency.
These exams include the General Scholastic Ability Test (GSAT),\footnote{\url{https://www.ceec.edu.tw/en/xmdoc/cont?xsmsid=0J180519944235388511}} Jitsuyo Eigo Gino Kentei (EIKEN),\footnote{\url{https://www.eiken.or.jp/}} the Advanced Subjects Test (AST),\footnote{\url{https://www.ceec.edu.tw/en/xmdoc/cont?xsmsid=0J180520414679660023}} and the Test of English Proficiency developed by Seoul National University (TEPS).\footnote{\url{https://en.teps.or.kr/about_teps.html}}
These indicate the critical importance of well-designed vocabulary questions in assessing language proficiency.
In this work, we seek to explore effective approaches for the automatic design of English vocabulary test items that accurately assess learners' vocabulary comprehension.
Automated test generation is expected to benefit both students and teachers. 
For students, it provides a greater variety of vocabulary contexts, helping them apply and understand words in different situations, which may contribute to improved retention. 
For teachers, it can reduce the burden of creating tests while maintaining consistent quality, allowing them to focus more on guiding student learning.

Vocabulary test questions typically follow two styles.
The first style, used in GSAT, ELKEN, and TEPS, presents a sentence with one word omitted and four options, only one of which fits the context.
For example, in the sentence ``Posters of the local rock band were displayed in store windows to promote the sale of their concert tickets,'' the target word ``concert'' is masked, and examinees must choose the correct word from distractors.
The second style, found in TOEFL iBT,\footnote{\url{https://www.ets.org/}} does not omit the target word but requires examinees to select the option matching its meaning.
As the second type provides a clear reference, distractor generation strategies may differ.
For instance, \cite{susanti2018automatic} consider the target word itself when generating distractors for the first type of exam.
Despite format differences, both styles rely on well-designed distractors to assess vocabulary knowledge.
This paper focuses on the first type, emphasizing the importance of generating distractors that test context-based inference skills.

The quality of a vocabulary test question is based on designing a stem that effectively tests the examinee’s understanding of a target word, followed by the creation of distractors that challenge the examinee to distinguish the target word from similar but incorrect options.
By presenting plausible yet incorrect options, effective distractors challenge students to discern subtle differences in meaning, ensuring that they do not merely guess but instead demonstrate a thorough comprehension of the vocabulary and the context in which it is used. 
This meticulous design ensures a more accurate assessment of the student's language proficiency and understanding.

Regarding the conditions required for a well-constructed vocabulary exam question, \cite{heaton1988writing} identifies several key elements for designing effective distractors:
(1)~the target word should have the same part of speech as the distractors,
(2)~the distractors and the target word should have a similar difficulty level,
(3)~the length of the target word and each distractor should be close,
(4)~and synonym pairs should be avoided in the options.
However, manual design of distractors is labor-intensive.
The automatic generation of distractors based on a given stem and target word, offering suggestions for teachers to select from, has gained attention in recent research. 
\cite{susanti2018automatic} propose extracting candidate distractors from lexical databases and word lists, measuring semantic relatedness for selection.
This method is limited by fixed word lists, reducing the diversity and flexibility of generated distractors.
\cite{liang-etal-2018-distractor} introduce a model trained on manual features to generate distractors resembling real exam questions.
\cite{chiang-etal-2022-cdgp} develop a framework combining neural networks and predefined rules to generate and rank diverse and effective distractors.
Previous methods~\citep{liang-etal-2018-distractor, chiang-etal-2022-cdgp} have achieved promising results in generating distractors for English vocabulary tests.
However, they require extensive training data, which is often infeasible for specialized vocabulary tests.
This highlights the need for effective approaches that do not rely on large-scale training data or predefined dictionaries.

Recent LLMs have demonstrated powerful semantic understanding and language generation abilities. 
Through in-context learning, combined with a few examples or specific prompting techniques like chain-of-thought prompting~\citep{wei-etal-2023-symbol,rubin-etal-2022-learning,liu2024let}, these models achieve excellent performance across various natural language processing tasks~\citep{brake-schaaf-2024-comparing,JALALI2024109801}, even without the need for additional training. 
Given these advantages, perhaps LLMs could be leveraged to address the aforementioned challenges.
However, it remains unclear whether LLMs can effectively perform the task of distractor generation.
This lead to the first research question (\textbf{RQ1}): 
\textbf{Can LLMs be directly utilized to generate distractors?}

Using LLMs to generate distractors directly could pose several challenges. 
First, although LLMs are proficient in language generation, producing effective distractors requires precise control over their similarity to the correct answer. 
Distractors must be sufficiently misleading but not too close to the correct answer. 
Additionally, distractors should not align too closely with the stem, which could result in multiple correct answers. 
Furthermore, distractors must be contextually appropriate and avoid any logical inconsistencies with the stem or other options. 
Another important consideration is the difficulty level of the distractors: they must strike a balance between being misleading and not overly simple or difficult. 
This balance is crucial to ensure the test effectively differentiates learners of varying proficiency levels.
Hence, we raise the second research question (\textbf{RQ2}): \textbf{How can we ensure that the generated distractors remain valid and do not introduce ambiguity or multiple correct answers?}

To address the research question mentioned above, this paper investigates the English vocabulary test items in the GSAT exam, one of Taiwan's most significant university entrance exams. 
The GSAT is widely recognized for its ability to assess students' vocabulary proficiency, making it an ideal subject for our study. 
Its test items encompass a broad range of vocabulary and include distractors of varying difficulty levels, offering a comprehensive framework for evaluating vocabulary mastery. 
By focusing on key features such as the range of vocabulary, the relationship between distractors and correct answers, and common error patterns among students, we seek to provide a detailed analysis of the effectiveness and challenges posed by existing test items. 
Given the direct impact of GSAT results on students' academic futures, our thorough examination of its vocabulary questions contributes not only to the refinement of vocabulary assessment methods but also to broader efforts in standardizing educational evaluation practices.

In this work, we propose the iterative selection with self-review (ISSR) framework, which assists teachers in designing English vocabulary exams by generating and validating distractors. 
The framework consists of three modules: a candidate generator, a distractor selector, and a distractor validator. 
We leverage a pretrained language model (PLM) to generate contextually relevant distractors, and introduce an LLM-based self-review mechanism to ensure the question remains valid with only one correct answer. 
This framework is expected to reduce the manual effort required to design distractors while potentially enhancing the diversity and accuracy of test items, making them more reflective of students' vocabulary comprehension.
Moreover, ISSR provides a flexible process, allowing for the integration of different advanced models into the framework. 
This adaptability enables the framework to evolve with advancements in language models, ensuring its versatility across various vocabulary exams and assessment needs. 
Additionally, ISSR does not rely on additional data for fine-tuning, which further enhances its efficiency and ease of use.
With the ability to adjust prompts, this approach could be adapted to a variety of vocabulary exams, making it a versatile tool for different assessment needs.
The details of each module are discussed in the following section.
In sum, our contributions in this paper are threefold:

\begin{itemize}
\item We investigate the challenge of using LLMs for automatic distractor generation in English vocabulary assessments, specifically focusing on the generation of contextually appropriate distractors that are valid and avoid ambiguity.
\item We explore the limitations of LLMs in this task, identifying key challenges and developing the ISSR framework based on these findings. 
\item Experimental results show that the distractors generated by ISSR perform well, and the proposed self-review mechanism effectively filters out invalid distractors.
\end{itemize}

\section{Related Work}
Recent research on automatic distractor generation typically divides the process into two steps~\citep{susanti2018automatic, ren2021knowledge, chiang-etal-2022-cdgp}:
(1)~generating a set of distractor candidates, 
and (2)~ranking these candidates to select the most plausible distractors.

\noindent \textbf{Distractor Candidate Generation.}
This step involves generating a broad set of distractor candidates to be further filtered in the next step.
\cite{susanti2018automatic} propose a method to obtain and rank distractors from lexical databases and dictionaries, specifically for the second type of exam described in Section~\ref{sec:intro}.
Their approach generates distractors based on the target word, stem, and correct answer, ensuring that distractors resemble the correct answer in meaning but remain distinguishable.
Candidates are retrieved from the text passage by focusing on words with the same part of speech and tense as well as sibling words from WordNet~\citep{fellbaum1998wordnet} and JACET8000~\citep{ishikawa2003jacet8000}, a dictionary tailored for Japanese English learners.
JACET8000 organizes 8,000 English words into eight levels of difficulty based on their frequency in the British National Corpus,\footnote{\url{http://www.natcorp.ox.ac.uk/}} supplemented by texts for Japanese students.
They observe that most distractors in human-created vocabulary questions share a similar difficulty level with the correct answer.
Based on these findings, they collect words matching the correct answer's difficulty level.
If too few candidates are identified, they expand the selection to include related terms from the Merriam-Webster Dictionary.\footnote{\url{https://www.merriam-webster.com/}}

Apart from vocabulary tests, open-domain multiple-choice questions assess knowledge in areas such as science, common sense, and trivia.
\cite{ren2021knowledge} generate distractors for such questions, noting that many options can be sourced from knowledge bases.
They propose selecting distractor candidates from WordNet and Probase~\citep{wu2012probase}.
One challenge they address is polysemy, where words like ``bank'' have multiple meanings.
To resolve this, they apply context-dependent conceptualization~\citep{kim2013context} using LDA~\citep{10.5555/944919.944937} to align the context with the most relevant concepts.
They calculate distractor scores based on a probability distribution of concepts derived from the target word and question stem.
The top $N$ scoring distractors are selected as the final candidates.

Previous work primarily utilized lexical databases and dictionaries to select distractor candidates.
\cite{chiang-etal-2022-cdgp} explore using pretrained language models (PLMs) for distractor generation, focusing on the CLOTH dataset~\citep{xie-etal-2018-large}, which contains teacher-created cloze-type questions.
Cloze-type questions present a paragraph with several words removed, requiring students to choose the best option for each blank.
These questions assess vocabulary knowledge and may also test grammar, such as prepositions.
Leveraging PLMs' ability to predict masked words or complete sentences, they fine-tune the models to generate plausible distractors for the question stem rather than correct answers.

\noindent \textbf{Distractor Candidate Scoring.}
In this step, candidates are ranked based on their suitability as effective distractors to select the most competitive options.
\cite{susanti2018automatic} emphasize that effective distractors should be semantically similar to the target word yet distinct from the correct answer, using a ranking formula based on semantic similarity and collocation with GloVe word vectors~\citep{pennington2014glove} and NLTK.
\cite{ren2021knowledge} advance the process by transforming candidates into 33-dimensional feature vectors to train ranking models, including AdaBoost~\citep{freund1997decision}, LambdaMART~\citep{burges2010ranknet}, and other rankers, enabling sophisticated selection strategies.
\cite{liang-etal-2018-distractor} propose NN-based and feature-based models for selecting plausible distractors, employing classifiers like logistic regression and random forests~\citep{breiman2001random} alongside adversarial training frameworks (IR-GAN)~\citep{10.1145/3077136.3080786} and cascaded learning~\citep{990517} for efficient filtering and ranking.
\cite{chiang-etal-2022-cdgp} introduce a scoring mechanism that combines PLM confidence scores with semantic similarity from contextual and word embeddings, assigning weights to scoring methods to identify the top $k$ distractors.
In summary, our work builds on prior methods for distractor generation and scoring but diverges by incorporating LLMs to address issues such as dictionary reliance, large training data needs, and filtering challenges.
By integrating LLMs, we seek to enhance flexibility and precision while retaining the strengths of earlier approaches.

\section{Data Analysis}\label{sec:analysis}
In this study, we selected questions from the GSAT English exam as our dataset, with past exam questions available on the College Entrance Examination Center (CEEC) website.\footnote{\url{https://www.ceec.edu.tw/xmfile?xsmsid=0J052424829869345634}}
We gathered exams from 2004 to 2016, comprising a total of 195 questions.
The GSAT English exam is designed to to assess whether candidates possess the fundamental academic abilities necessary for university education, serving as a preliminary screening criterion for university admissions and program selection.
We consider these carefully curated exam questions to be a suitable benchmark for exploring methods that support teachers in evaluating Taiwanese high school students' second language learning abilities. 
Although the dataset is specific to Taiwan, the insights gained from this study may have broader applicability in similar educational contexts.

To develop a method for generating English vocabulary questions that align with teacher-designed tests and to examine distractor characteristics that effectively mislead students, we analyze question features in the following section.
First, as vocabulary selection is a critical component of vocabulary tests, we analyze the range of words chosen and the methods used to evaluate their difficulty.
Next, we examine the relationship between target words and distractors, focusing on the characteristics of distractors in questions with low pass rates.
Finally, we analyze factors influencing students' selection of distractors, identifying key elements that disperse their choices.

\subsection{Analysis of CEEC wordlist}
\label{sec: voc_source}
The CEEC has published a Senior High School English Wordlist,\footnote{\url{https://www.ceec.edu.tw/SourceUse/ce37/ce37.htm}} detailing the vocabulary that Taiwanese high school students should understand before taking the GSAT and AST English tests.
Each word in the list includes its difficulty level and part of speech, covering categories such as nouns, verbs, adjectives, adverbs, articles, pronouns, conjunctions, and prepositions.
Adverbs formed by adding ``-ly'' to adjectives are not listed separately.
The vocabulary is classified into six levels of difficulty, primarily based on word frequency from the Cobuild English Dictionary~\citep{2302} and the Cobuild English Dictionary for Advanced Learners~\citep{17146}, with additional rules applied for refinement.

\noindent \textbf{Derivatives}: As derivatives are typically considered more difficult than their root words, in most cases, the difficulty of a derivative is higher than that of its root.
For example, the word ``true'' is assigned a difficulty level of 1, its derivative ``truth'' is level 2, and ``truthful,'' derived from ``truth,'' is classified at level 3.
However, if a derivative is more common than its root, then the derivative may be less difficult than the root.
For example, as ``consideration'' is regarded as higher-frequency than its root word ``considerate'', ``consideration'' is level 3 whereas ``considerate'' is level 5.

\begin{figure}[t]
    \centering
    \includegraphics[width=10cm]{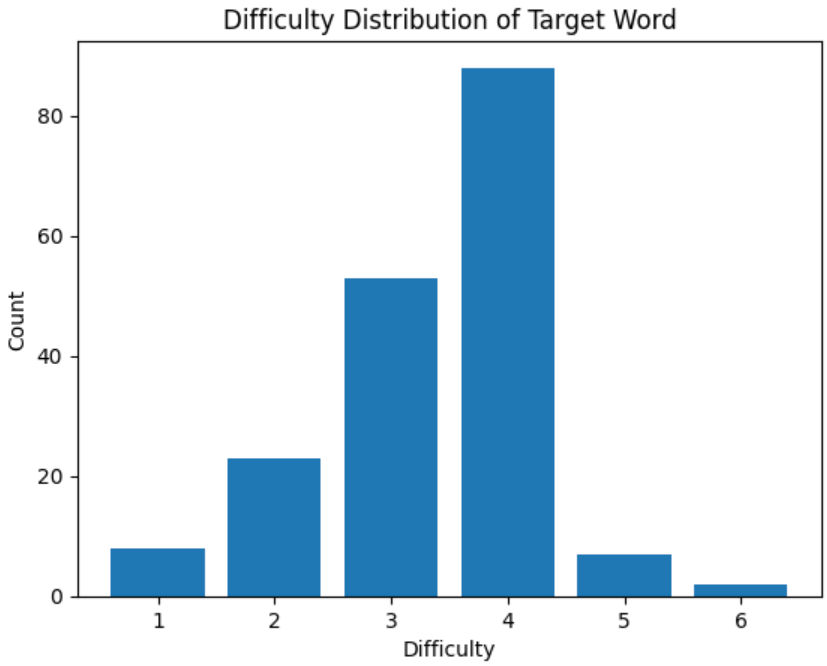}
    \caption{Difficulty distribution of target word}
    \label{tab:freq}
\end{figure}

\noindent \textbf{Analyzing GSAT Exam Vocabulary Levels}: 
To examine the difficulty distribution of vocabulary used in the GSAT English exam, we analyzed the target words' difficulty levels by considering the word difficulty of each question's target word.\footnote{Since this exam is designed for Taiwanese students, a few low-frequency words from Chinese culture in the Cobuild English Dictionary are included (e.g., ``bamboo'', ``chopsticks'', ``dumpling''). }
Because some target words appear in different forms (e.g., due to tense changes), we lemmatized them using the spaCy\footnote{\url{https://spacy.io/}} toolkit to ensure their base forms matched the entries in the Senior High School English Wordlist.  
Figure~\ref{tab:freq} illustrates the difficulty distribution of target words in the GSAT exam, with the x-axis showing difficulty levels and the y-axis indicating the number of words at each level.  
The majority of words fall into difficulty levels 3 and 4, with approximately 60 and 80 words, respectively, whereas fewer words are distributed across levels 1, 2, 5, and 6.  
This indicates that the GSAT English exam primarily tests words of moderate difficulty, with only a few being particularly easy or difficult.  
Table~\ref{tab: word_difficulty_level_example} provides examples of words from difficulty levels 1 to 6.  

\begin{table}[t]
\centering
\small
\begin{tabular}{cccccc}\toprule
Level 1 & Level 2 & Level 3 & Level 4 & Level 5  & Level 6 \\ \midrule
alone & beef & bandage  & ashamed & considerate & anonymous \\
bathroom & arrange & crown  & aluminum & astronaut & contradiction \\
already & calendar & deposit& container & bruise & eloquence \\
eagle & fever & envy& emphasis &  contemporary & humanitarian \\
always& interview & fur&inflation & immense & prosecution \\
\bottomrule
\end{tabular}
\caption{Vocabulary of various difficulty levels}\label{tab: word_difficulty_level_example}
\end{table}

\subsection{Relation between distractors and answer}\label{sec:distractor_analysis}
\noindent \textbf{Cosine similarity between distractors and answer}: 
Cosine similarity is a commonly used metric for analyzing word similarity~\citep{chiang-etal-2022-cdgp,susanti2018automatic,jiang-lee-2017-distractor}: the higher the similarity between two words, the more likely they are to share the same semantic meaning.  

One intuitive approach in vocabulary tests is to design distractors that are semantically similar to the target word.
However, if the similarity is too high, the question may become invalid, as both the target word and the distractor could be correct answers within the context.
We measure the cosine similarity between the target word and distractors to determine the appropriate level of similarity for a plausible choice.
Using the \texttt{en\_core\_web\_lg}\footnote{\url{https://spacy.io/models/en}} model from spaCy, we convert words into vectors and compute their cosine similarity.
The average cosine similarity of distractors in the GSAT English dataset is 0.290, with a standard deviation of 0.131, suggesting that distractors exhibit low to moderate similarity, avoiding invalidation of the question.

To investigate whether questions with low pass rates exhibit low cosine similarity, we analyzed questions with pass rates below 60\% and calculated the cosine similarity between the target word and its distractors
The results show that the average cosine similarity in these questions is 0.257, with a standard deviation of 0.08.
Our analysis indicates that high-error-rate questions do not exhibit significantly higher cosine similarity between the target word and distractors.
This suggests that high semantic relatedness is not the primary factor misleading students.
Determining an optimal level of cosine similarity remains inconclusive, as high similarity may invalidate the question by providing multiple correct answers.
Thus, we chose at this stage not to incorporate cosine similarity between the target word and distractors into our framework.

\noindent \textbf{Difficulty Difference Between Distractors and Answers}:
In Section~\ref{sec: voc_source}, we analyzed the difficulty distribution of vocabulary selected by GSAT as target words.
In this section, we further examine the difficulty levels of words that teachers choose as distractors to assess students' English vocabulary skills.
As noted by \cite{heaton1988writing}, the relative difficulty of distractors compared to the correct answers is critical to question quality.
If distractors are significantly harder or easier than the correct answer, they are less effective in misleading students, making it harder to evaluate their true understanding of the target word's usage.

\begin{figure}[t]
    \centering
    \includegraphics[width=10cm]{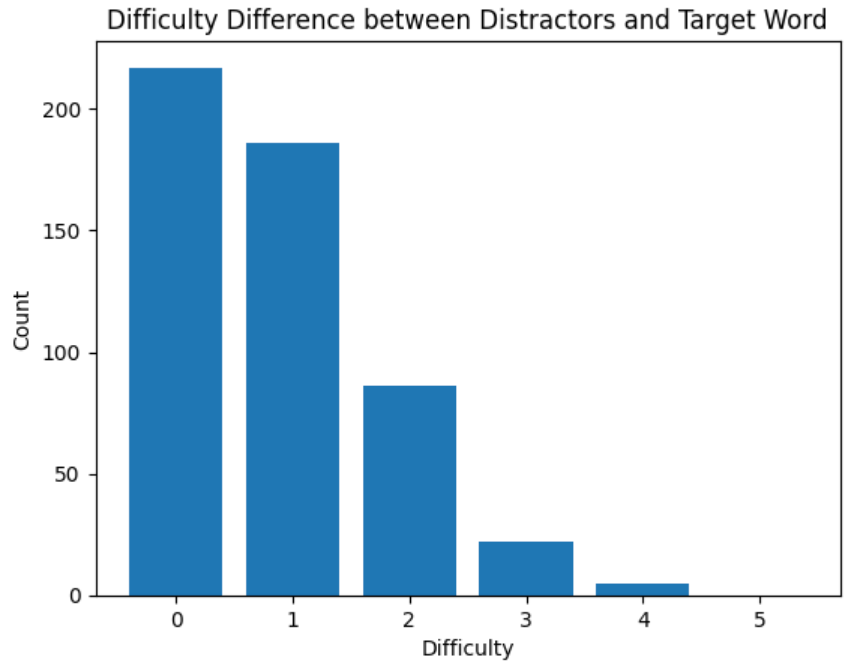}
    \caption{Difficulty difference between distractors and answers}
    \label{fig:dif_dif}
\end{figure}

We analyze the difficulty difference between the target word and the distractors for each exam question. 
Similar to the method used in Section~\ref{sec: voc_source} for calculating the difficulty of target words, we first lemmatized the distractors using the \texttt{en\_core\_web\_sm} model in the spaCy toolkit. 
We then determined their difficulty levels based on the Senior High School English Wordlist.
After that, we compared the difficulty differences between the target words and their corresponding distractors.
Figure~\ref{fig:dif_dif} shows the difference between the distractors and the target word.
The x-axis represents the difficulty levels of the target words, and the y-axis represents the number of words at each difficulty level. 
As shown in Figure~\ref{fig:dif_dif}, the difficulty difference between the target word and distractors in most questions falls within the range of 0 to 2 levels.
Only a few questions exhibit a difficulty difference of 3 or 4 levels between the target word and the distractors.

We found that most questions keep the difficulty level of distractors within one level higher than that of the answer.
Since difficulty levels are classified based on word frequency, words with similar frequency levels are more likely to be grouped together in exam questions.
To ensure that the generated distractors align with the nature of the analyzed exam questions, our framework incorporates the difficulty difference between the distractor and the answer.

\subsection{Analysis of question types in which students commonly make mistakes}


The Senior High School English Wordlist classifies words into difficulty levels based on frequency of occurrence.  
However, it is unclear whether lower-frequency words present greater challenges for students.  
To investigate, we collected the selection rates for each option as published by CEEC, where the \textit{selection rate} indicates the proportion of students choosing each option.  
The \textit{pass rate} specifically represents the selection rate of the correct option, reflecting the percentage of students who answered correctly.  
In this section, we analyze the relationship between pass rates and answer difficulty, as defined by the classifications in the Senior High School English Wordlist.  

\begin{figure}
    \centering
    \includegraphics[width=10cm]{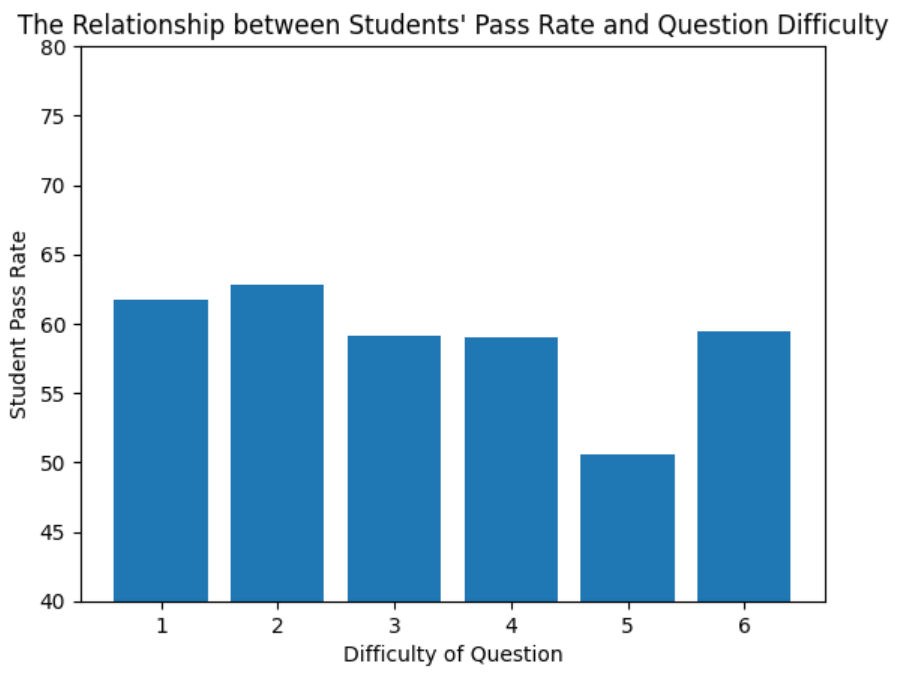}
    \caption{Relationship between student pass rates and question difficulty}
    \label{fig:dis_rate}
\end{figure}

Figure~\ref{fig:dis_rate} shows the pass rates of students for questions across six difficulty levels.
The Pearson correlation between question difficulty and student pass rate is -0.118, with a $p$-value of 0.113.
This suggests that there is no significant correlation between overall pass rate and question difficulty.
Based on our analysis and the exam statistics from the past few years, vocabulary words across all difficulty levels appear to be effective as test items for assessing students' mastery of English vocabulary.

At this stage, we have not developed specific methods for designing questions that focus on polysemous words.\footnote{We conducted an in-depth analysis of how polysemous words and their less common meanings, as interpreted through translation tools, pose unique challenges for Asian students learning English; detailed methodology and findings are provided in~\ref{sec:translation}.}
Nonetheless, these preliminary findings suggest that further exploration is warranted regarding the role of polysemous words in exams and their impact on student performance, which we plan to address in future work.

\section{Framework of Iterative Selection with Self-Review}
\begin{figure*}
    \centering\includegraphics[width=1\linewidth]{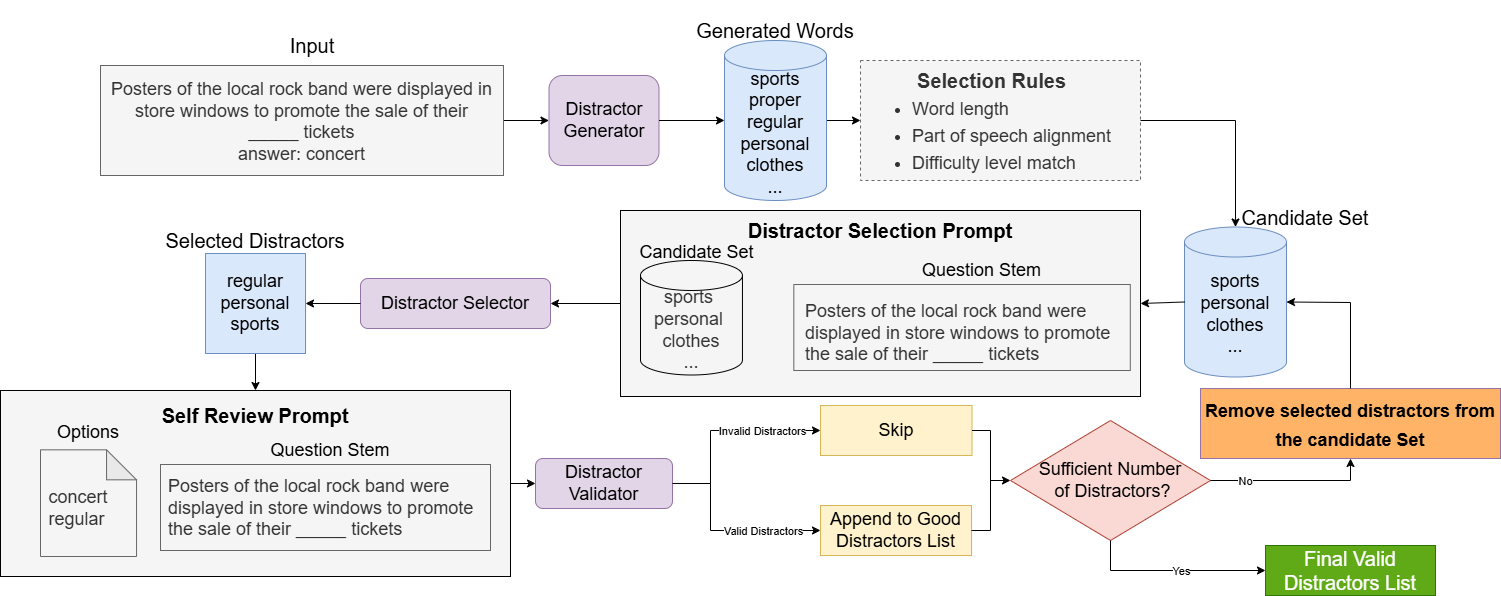}
    \caption{ISSR framework}
    \label{fig:ISSR_framework}
\end{figure*}

In this work, we propose the ISSR framework, which generates a set of distractors to assist teachers in designing English vocabulary tests.
As shown in Figure~\ref{fig:ISSR_framework}, the framework is composed of a candidate generator, a distractor selector, and a distractor validator.

\subsection{Candidate generator}
\label{sec:candidate_generator}

Given a question stem and the target word, this module generates a large number of candidate distractors for further selection.  
To ensure the distractors are confusing yet plausible within the context, without being easily dismissed, we adapted a pretrained language model (PLM) as the distractor generator.  
We chose a PLM over lexical databases commonly used in previous work for several reasons.  
(1)~A PLM, designed to generate words within a sequence by understanding context, produces coherent and semantically meaningful outputs, making it highly effective for generating contextually relevant distractors.  
(2)~Unlike predefined rules that rely on lexical databases such as WordNet and Probase, PLMs generate a significantly larger quantity of distractor candidates.  
We leveraged the BERT-base model named CDGP-CSG, trained on the CLOTH dataset proposed by \cite{chiang-etal-2022-cdgp}.
Although CDGP-CSG was originally designed for a different context, it generates distractors that appear plausible in vocabulary exams.
The target word in the question stem is masked and input into the BERT model for fill-in-the-blank processing.

The generated results undergo additional conditional filtering to ensure their suitability based on Section~\ref{sec:analysis} and the criteria for effective distractors outlined by \cite{heaton1988writing}.

The filtering rules are as follows:
\begin{itemize}
\item The length difference between the answer and distractor should not exceed two characters.
\item The answer and distractor must share the same part of speech.
\item The difficulty levels of the answer and distractor should be closely matched.
\end{itemize}

To verify whether the answer and the distractors share the same part of speech, we insert the generated distractors back into the stem and utilize the NLTK toolkit to analyze their parts of speech.
Additionally, to ensure that the difficulty levels of the answer and distractors are comparable, we reference the vocabulary list provided by the Taiwan CEEC. 
After lemmatizing both the answer and the distractors, we retain only those distractors with a difficulty difference of zero or one.
This process is designed to extract up to 50 distractors for the candidate pool.

\subsection{Distractor selector}

Due to the ineffectiveness of directly generating distractors using the LLM, we instead present a pool of candidate distractors from which the LLM is used to select the most suitable ones.  
Further discussion on the limitations of direct distractor generation will be provided in the following section.  
In this step, we input the candidate pool into the prompt and instruct the LLM to choose the top $k$ most appropriate distractors.  
After generating distractors using the candidate generator, post-processing filters out those that are unsuitable for English vocabulary exams.  
Given a question stem, target word, and a set of distractor candidates, we develop a distractor selector to select $k$ appropriate distractors.  
Previous research primarily employs ranking models~\citep{liang-etal-2018-distractor} or manually defined formulas~\citep{susanti2018automatic,chiang-etal-2022-cdgp}.  
Although ranking models identify appropriate distractors, they require extensive training data and do not necessarily generalize well across different exam scopes.  
The ranked words may align with specific difficulty levels but lack applicability across varying exam requirements.  
Manually defined formulas are also suboptimal, as it is challenging to define the ideal degree of semantic similarity between a distractor and the target word.  
Distractors with high semantic similarity to the target word risk invalidating the question by creating multiple correct answers.  

According to our analysis in Section~\ref{sec:distractor_analysis} using GSAT English exam data, there is no significant correlation between the cosine similarity of distractors, suggesting that semantic similarity may not be the primary factor in challenging examinees.  
To address this, we integrate an LLM into the framework for generating English vocabulary questions.  
Leveraging the LLM's semantic understanding and natural language generation capabilities, we utilize in-context learning to select appropriate distractors based on instructions without requiring additional training.  
An LLM-based distractor selector is incorporated into ISSR to select the top $k$ distractors from the candidate set.

\subsection{Distractor validator}
Sometimes, the candidate distractors generated in the previous steps could also serve as plausible answers, potentially resulting in multiple correct answers in the question.  
As the test design requires only one correct answer among the four options, further filtering ensures that distractors cannot function as valid correct answers.  
To address this, we propose a self-review mechanism to assess the validity of each distractor.  
New questions are formulated by pairing each distractor with the correct answer as a binary choice, allowing the LLM to evaluate each distractor's suitability.  
If the LLM selects the distractor as the correct answer, we take this to indicate that the distractor is unsuitable, as it yields a question with multiple valid answers.  
We also explore additional self-review methods, with experimental results discussed in the next section.  
If fewer than $k$ qualified distractors remain, valid and invalid distractors are removed from the candidate set, and the distractor selector is invoked to choose additional appropriate distractors.

\section{Experiments}
\subsection{Experimental setup}\label{sec:setup}

In the experiment, we utilized English vocabulary questions from the GSAT spanning from 2006 to 2018.
The dataset consisted of 195 questions. 
We applied both zero-shot and few-shot prompting approaches in our experiments.  
For few-shot prompting, the first two questions were used as demonstrations for in-context learning, and the remaining 193 questions served as the test set. 
In the zero-shot prompting setting, the same 193 questions were directly used as the test set without prior demonstrations. 
To ensure the LLM provided the desired information, such as the selected distractors, we structured the prompts with specific formatting instructions, guiding the LLM to follow a designated response format.\footnote{The prompts used in this work are presented in \ref{sec:prompts}.}  
The temperature for all LLMs was set to 0.7.  
To verify that our ISSR framework improves the automatic distractor generation capabilities of LLMs, we compared the ISSR framework against the following baselines: CDGP, GPT-3.5~\citep{ouyang2022training}, and GPT-4o-mini~\citep{islam2024gpt}. 
We followed the original CDGP setup as specified, without any modifications. 
The GPT-3.5 model used in this comparison was \texttt{gpt-3.5-turbo-0125}, and GPT-4o-mini used in this comparison was \texttt{gpt-4o-mini-2024-07-18}.




\subsection{Experimental results}
\label{sec:five_two}

\begin{table}[t]
\centering
\small
\begin{tabular}{lrrrrrrr}\toprule
Method &F1@3 &F1@10 &NDCG@3 &NDCG@10 &NDCG@30 \\\midrule
CDGP &0.51\% &1.10\% &1.19\% &3.15\% &5.95\% \\
GPT-3.5 w/ zero-shot &0\% &0.32\% &0\% &0.59\% &1.34\% \\
GPT-3.5 w/ few-shot &0.35\% &0.56\% &0.85\% &1.71\% &3.27\% \\
Direct selection & 0.35\% & 0.24\%&0.68\%&0.85\%&1.40\% \\ \hline
ISSR &\textbf{1.55\%} &\textbf{2.07\%} &\textbf{3.57\%} &6.31\% &\textbf{9.82\%} \\
\hspace{5mm} w/o self-review &1.04\% &1.91\% &3.11\% & \textbf{6.78\%} &7.44\% \\
\bottomrule
\end{tabular}
\caption{Distractor generation performance}\label{tab: main_result}
\end{table}

Table~\ref{tab: main_result} compares the results of ISSR to that of other models.
Since the primary goal of this work is to generate a sufficient number of high-quality distractors for teachers to select from, we extended the generated distractor count to 30.
F-score and NDCG were adopted as the evaluation metrics.
In addition to the baselines mentioned in Section~\ref{sec:setup}, we assessed the LLM's ability to select suitable distractors from the entire vocabulary, referred to as the ``Direct selection'' method.
Specifically, we instructed GPT-4o-mini to select distractors directly from the vocabulary list provided by the Taiwan CEEC.
The choice of GPT-4o-mini as the selector was based on its reputation as a stronger LLM than GPT-3.5, making it an ideal model for evaluating distractor selection due to its advanced language processing capabilities.
If the LLM performs well under this approach, it would suggest that LLMs have sufficient capability to handle wide-ranging distractor selection autonomously.
As shown in Table~\ref{tab: main_result}, ISSR not only surpasses the direct use of GPT-3.5 for distractor generation and direct selection with GPT-4o-mini, but also outperforms CDGP.
This finding suggests that a well-designed distractor candidate set plays a crucial role in generating high-quality distractors, emphasizing the importance of a well-curated candidate generation process.
The findings indicate that ISSR is more effective in generating distractors that closely align with those found in actual exams, thereby offering educators a broader selection of viable distractor options.
Furthermore, Table~\ref{tab: main_result} shows that the integration of a self-review mechanism, which removes distractors that could potentially invalidate the question, contributed to an overall improvement in performance.

Note that the performance of all methods is lower because the models select distractors from the output of the distractor generator, which may not include suitable distractors.
However, without filtering candidates through the distractor generator, the distractor selector would need to choose from thousands of candidates.
In Section~\ref{sec:distractor_select}, we evaluated the performance of selecting distractors from a limited candidate pool consisting of teacher-designed distractors.
Further discussion and examples comparing ISSR to direct generation methods can be found in \ref{sec:appendix_direct_inference}.

An interesting finding is that both zero-shot prompting and few-shot prompting using GPT-3.5 for distractor generation are outperformed by CDGP on this task.
This is likely due to the inherent instability of GPT-3.5 during generation.
Our experiments show that when GPT-3.5 is tasked with generating a large set of distractors in a single round, it often produces repetitive content, as shown in Table~\ref{tab:appendix_example_large_count_single_round}.
To address this, we experimented with generating a smaller number of distractors over multiple rounds, which partially reduced repetition.
However, this approach requires modifying the prompt after each round to prevent the model from producing similar outputs, as GPT-3.5 tends to generate redundant content if the prompt remains unchanged.

In our implementation, we addressed this by adding explicit instructions to the prompt, directing the LLM to avoid repeating previously generated distractors, as demonstrated in Table~\ref{tab:appendix_example_small_count_per_round}.
However, as restrictions on the reuse of prior outputs increase, GPT-3.5’s generation can become increasingly erratic, resulting in repetitive or less relevant content.
This limitation poses a significant challenge when a large number of unique distractors are required.
This finding demonstrates the advantages of the ISSR framework, which mitigates these issues by reframing the task for the LLM: instead of directly generating distractors, ISSR focuses on selecting suitable distractors from a pre-generated candidate set.
This approach leverages the LLM’s semantic capabilities to ensure that the selected distractors are both diverse and contextually appropriate.

\noindent \textbf{Performance of different candidate generators.}
The ISSR framework is highly flexible, allowing for the integration of various models as candidate generators, which are used to produce candidate distractors.
In this experiment, we tested the impact when selecting different sources as the candidate generator on ISSR.
The comparison involves three settings: 
(1)~no candidate generator, denoted as None, where distractors are directly generated using GPT-3.5 with zero-shot prompting and a self-review process is applied to filter out invalid distractors; 
(2)~a standard BERT-base model; 
and (3)~a BERT-base model fine-tuned on the CLOTH dataset, referred to as the CDGP-CSG model. 
Note that the ISSR framework incorporates predefined rules, as outlined in Section~\ref{sec:candidate_generator}, to filter candidates generated by the candidate generator. 
For consistency and fairness, the same filtering mechanism was applied uniformly across all candidate generators.

Table~\ref{tab: main_candidate_generator} presents the results when using different models as the candidate generators. 
The results show that CDGP-CSG outperforms BERT-base.
In contrast, the standard BERT-based candidate generator exhibits weaker performance.
Since BERT was trained on masked token prediction and next sentence prediction, it is intuitive that BERT would attempt to generate tokens that fit the given contexts, leading to the generation of suboptimal distractor candidates.
The CDGP-CSG model, on the other hand, uses a BERT model that is fine-tuned to generate distractors on the CLOTH dataset.
Although the CLOTH dataset differs from the GSAT in terms of exam scope, CDGP-CSG has learned to generate plausible distractors for a given question stem instead of merely generating words that fit the contexts.

\begin{table}[t]
\centering
\small
\setlength\tabcolsep{2pt}
\begin{tabular}{lrrrrrr}\toprule
Candidate generator &F1@3 &F1@10 &NDCG@3 &NDCG@10 &NDCG@30 \\\midrule
None &0.35\% &0.56\% &0.78\% &1.56\% &2.46\% \\
BERT-base-uncased &0.52\% &1.12\% &0.85\% &2.69\% &5.71\% \\
CDGP-CSG &\textbf{1.55\%} &\textbf{2.07\%} &\textbf{3.57\%} &\textbf{6.31\%} &\textbf{9.82\%} \\
\bottomrule
\end{tabular}
\caption{Performance using different candidate generators}\label{tab: main_candidate_generator}
\end{table}

\noindent \textbf{Results Using Different LLMs for Distractor Selection.} 
The pre-training datasets and architectures of different LLMs vary, leading to differences in their abilities to select distractors.
In this experiment, we assessed the capabilities of different LLMs in selecting distractors. 
The comparison involved three different models with varying parameter sizes: (1)~GPT-3.5, (2)~\texttt{Llama3-8B}~\cite{dubey2024llama}, and (3)~\texttt{Llama3-70B}.
For each model, both few-shot and zero-shot settings were evaluated; the CDGP-CSG model was used as the candidate generator.

Table~\ref{tab: main_llm_selection} shows the results using different LLMs to select distractors:
GPT-3.5 with zero-shot prompting outperforms both \texttt{Llama3-8B} and \texttt{Llama3-70B}, despite the significant difference in their parameter sizes.
Notably, \texttt{Llama3-8B} exhibits performance comparable to \texttt{Llama3-70B}, suggesting that in this task, model size does not necessarily correlate with improved performance.
Additionally, Table~\ref{tab: main_llm_selection} indicates that the distinction between few-shot and zero-shot settings has minimal effect on the performance of LLMs when selecting the most appropriate distractors from a candidate set.
The zero-shot slightly outperforms few-shot, suggesting that the LLM is capable of effectively selecting distractors without requiring prior demonstrations.
Based on these findings, we adopt GPT-3.5 with zero-shot prompting in the ISSR framework.

\begin{table}[t]
\centering
\small
\setlength\tabcolsep{3.5pt}
\begin{tabular}{lrrrrrr}\toprule
Model &F1@3 &F1@10 &NDCG@3 &NDCG@10 &NDCG@30 \\\midrule
Llama3 8B w/ Zero-Shot &0.52\% &1.36\% &1.36\% &3.60\% &7.47\% \\
Llama3 8B w/ Few-Shot &0.51\% &1.12\% &1.55\% &3.32\% &7.06\% \\
Llama3 70B w/ Zero-Shot &0.69\% &0.96\% &1.62\% &3.20\% &7.68\% \\
Llama3 70B w/ Few-Shot &0.86\% &1.43\% &1.95\% &4.01\% &8.49\% \\
GPT-3.5 w/ Zero-Shot &\textbf{1.55\%} &\textbf{2.07\%} &\textbf{3.57\%} &\textbf{6.31\%} &\textbf{9.82\%} \\
GPT-3.5 w/ Few-Shot &\textbf{1.55\%} &1.59\% &3.05\% &4.99\% &8.61\% \\
\bottomrule
\end{tabular}
\caption{Results using different LLMs for distractor selection}
\label{tab: main_llm_selection}
\end{table}

\section{Analysis and Discussion}

In this section, we further analyze the impact of the proposed framework.  
A human evaluation is also conducted, as detailed in \ref{sec:human_eval}.  

\subsection{Impact of candidate set size}
\label{sec:candidate_set_size}

Perhaps it is natural to think that providing more distractor candidates to the distractor selector would improve the outcome, as it gives the LLM more options to choose from.
However, it remains unclear whether the LLM can accurately select proper distractors from a large candidate set.
During the prompt design phase, we found that the size of the candidate set can influence the quality of the LLM's output, which ultimately affects the performance of the distractor selector.
As the size of the candidate set increases, the likelihood of the LLM generating distractors that do not appear in the distractor candidate set also increases.

To more precisely analyze the impact of the candidate set size on the LLM's ability to effectively select distractors, we conducted an experiment to determine the optimal candidate set size.
In this experiment, we utilized ISSR and varied the number of candidate distractors provided to the distractor selector.
Specifically, we extracted the stem and target word from the original teacher-designed test questions and generated multiple prompts, each corresponding to a different candidate set size generated by the distractor generator.
We then evaluated whether the distractor selector could accurately select distractors from within the provided candidate set.
The LLM used in ISSR was GPT-3.5.

Table~\ref{tab: llm_good_select_rate} shows that the success rate of selecting three different distractors varies with different sizes of candidate sets.
The success rate reflects how accurately the distractor selector identifies teacher-designed distractors from the candidates. 
Specifically, to evaluate the LLM's selection capability, we incorporated the original teacher-designed distractors into the candidate pool and supplemented them with an additional $n$ random candidates.
As shown in Table~\ref{tab: llm_good_select_rate}, we found a negative correlation between the size of the candidate set and the rate at which the LLM accurately selects distractors.
Although the LLM reliably selects distractors from a candidate set of size 50, its performance becomes more unstable as the candidate set size increases, ultimately resulting in a drop in accuracy to 90.67\% when the candidate set size reaches 300.
We hypothesize that the LLM's responses may not align with the original prompt's request to select suitable words from the distractors because a larger candidate set may disrupt the LLM's understanding that the current context still pertains to the candidate set.
This could lead to the LLM inferring relationships between distractors within the set, ultimately affecting the outcome.
We conclude that reducing the candidate set size is necessary to prevent the LLM's output from being disrupted, while still ensuring there are enough distractors available for selection.
Ultimately, we set the candidate size to 50. 

\begin{table}[t]
\centering
\begin{tabular}{lrr}\toprule
Candidate size & Selection rate\\\midrule
300 &90.67\% \\
200 &92.75\% \\
100 &97.93\% \\
50 &\textbf{98.79\%} \\
\bottomrule
\end{tabular}
\caption{Relationship between candidate size and successful distractor selection rate}\label{tab: llm_good_select_rate}
\end{table}

\subsection{Impact of different self-review methods}
\label{sec: different_self_review}
The main goal of a distractor validator is to ensure that the generated distractors do not serve as valid correct answers. 
Since LLMs can solve a wide range of problems, there are numerous approaches involving LLM inference to achieve this goal.\footnote{We tested the performance of LLMs on English vocabulary questions, as detailed in \ref{sec:LLM_answer_ability}.}
To investigate the LLM's ability to distinguish between valid and invalid distractors using different queries, we designed an experiment in which we examine the effectiveness of three different self-review methods with different prompts.
We used GPT-3.5 as the distractor validator and utilized 193 GSAT English exam questions for evaluation.
The core method we used to validate the appropriateness of a distractor involves presenting both the distractor and the question to the LLM, allowing it to evaluate its suitability using various assessment approaches. 
Ultimately, we selected the most effective evaluation method as the final self-review approach.

We separated the three golden distractors in each question into individual queries, resulting in a total of 579 queries for the LLM to evaluate the suitability of each distractor.
By providing the golden distractor and target word, the distractor validator is expected to respond by recognizing the golden distractor as a valid option.
The three prompts have the following query objectives:

\noindent \textbf{Independent Suitability Judgment.} 
In this prompt, we first instructed the LLM by informing it that it is an English teacher designing a vocabulary test for students. 
After providing the stem and target word, we asked the LLM whether the generated distractors were appropriate for this question.
This design tested whether the LLM can independently judge whether a distractor is suitable for the given question.
The results indicate that the LLM confidently identified the golden distractor as a suitable option only 4 out of 579 times.

\noindent \textbf{Semantic Consistency Check.} 
In this version, we created two sentences by filling the incomplete stem with either the target word or the distractor.
We first informed the LLM that the sentences differ by only one word,
after which we asked whether they convey the same meaning.
This design tested whether the LLM could discern whether the meaning changed when the target word was replaced with the distractor, thereby determining if the distractor had altered the original meaning.

Since distractors typically convey a meaning different from the correct answer, especially when inserted into the stem, a valid distractor should result in a sentence with a meaning different from that when the correct answer is used.
The results show that in 397 out of 579 instances, the LLM identified that the golden distractor changed the meaning of the sentence.
This approach was thus more effective than the previous method, where the LLM was asked to directly assess the suitability of the distractors. 
This improvement may stem from the more explicit evaluation of meaning shifts, rather than relying solely on the model’s inherent understanding of distractor quality. 
Additionally, the results suggest that the LLM lacks a clear internal understanding of the characteristics a distractor should possess, highlighting the need for more guided context-based evaluation.

\noindent \textbf{Binary Choice Validation.} 
According to the results presented in \ref{sec:LLM_answer_ability}, we found that LLMs excel at solving vocabulary test questions.
Therefore, in this version, we formulate the task of evaluating distractor suitability as a binary choice.
Specifically, the LLM is presented with both the correct answer and a distractor and is tasked with selecting the correct answer to fill in the stem, thereby determining the appropriateness of the distractor.
Note that, in contrast to standard single-choice vocabulary questions, the prompt allowed the LLM to consider both two options (i.e., the target word and the distractor) as possible answers to be filled in the stem, enabling it to recognize cases where either option could plausibly complete the question stem.
The LLM selecting the distractor as a correct answer indicated that the distractor could potentially serve as an alternative valid answer when inserted into the stem. 
However, this is problematic because distractors are specifically designed to be incorrect, meaning they should never be considered valid answers when filling the stem.
Hence, the intended outcome is that the model consistently identifies the target word as the correct answer from the two options and uses it to fill in the stem.
The results show that the LLM correctly selected the target word as the answer in 563 out of 579 instances.
This indicates that the method is effective in filtering out unsuitable distractors, as it forces the model to make a clear distinction between the correct answer and the distractor. 
By placing the distractor in direct comparison with the target word, the model better assesses the appropriateness of each option, reducing the likelihood of selecting distractors that could be mistakenly viewed as fitting answers for the question, which would undermine the question’s integrity.
Among the 16 questions answered incorrectly, only two instances involved the LLM responding with ``Both are good.''
This suggests that the LLM tends to select a single option as its output rather than indicating that both options are equally acceptable.
Since binary choice validation performed the best, we adopted this approach as the self-review mechanism in ISSR.
Detailed prompts used in this experiment can be found in Table~\ref{tab:prompt_self_review}.

\subsection{Distractor selection evaluation}
\label{sec:distractor_select}
As discussed in Section~\ref{sec:five_two}, direct distractor generation using LLMs may not be the optimal solution due to their instability and limitations in generating a large number of distractors. 
However, it remains unclear whether LLMs are able to select appropriate distractors from a pool of distractor candidates.
To this end, we examine the effectiveness of LLMs in selecting plausible distractors from a given candidate set. 
First, for each question stem and target word from the GSAT English vocabulary test, we generated 10 distractor candidates using the \texttt{BERT-base-uncased} model as the candidate generator. 
Next, we randomly replaced some of these candidates with the golden distractors from the actual exam, ensuring that each generated candidate set included appropriate distractors.
We then provided the stem, target word, and candidate set to various LLMs, asking them to select the top 3 most suitable distractors. 
Table~\ref{tab: LLM_selection_ability} shows the results selected by the LLMs, indicating that most LLMs selected appropriate distractors from a well-curated candidate set.
However, when distractors were not drawn from such specially designed candidate sets, the F-score drops significantly (presented in Table~\ref{tab: main_result}), indicating that the bottleneck lies in generating suitable candidate distractor sets.
Therefore, improving the generation of distractor candidate sets is key to enhancing the overall system performance.

\begin{table}[t]\centering
\small
\begin{tabular}{lrrrr}\toprule
Model &Candidate set size &F1@3 &NDCG@3 \\\midrule
GPT-3.5 &10 &35.58\% &58.60\% \\
Llama3 8B &10 &43.52\% &71.15\% \\
Llama3 70B &10 &33.85\% &56.65\% \\
\bottomrule
\end{tabular}
\caption{LLM distractor selection abilities}\label{tab: LLM_selection_ability}
\end{table}

\begin{table}[t]\centering
\small
\begin{tabular}{crrrrrr}\toprule
Selection size &F1@3 &F1@10 &NDCG@3 &NDCG@10 &NDCG@30 \\\midrule
30 &0.52\% &1.19\% &1.30\% &3.20\% &6.44\% \\
10 &0.69\% &0.96\% &1.55\% &3.03\% &7.49\% \\
3 &\textbf{1.55\%} &\textbf{2.07\%} &\textbf{3.57\%} &\textbf{6.31\%} &\textbf{9.82\%} \\
\bottomrule
\end{tabular}
\caption{Performance of ISSR under different distractor selection count per round}\label{tab: distractors_count_per_round}
\end{table}

\section{Impact of Distractor Selection Count on LLM Performance}

In our experimental results, we observed that LLMs are able to select appropriate distractors from a candidate set. 
However, we also found that the number of distractors requested from the LLM during each selection round not only affects the quality of the chosen distractors but also impacts the overall validity of the results.
In this subsection, we explore how varying the number of distractors selected per round influences the LLM's ability to effectively choose suitable distractors and maintain consistency in the selection process.

In the experimental setup, the LLM was to select 30 distractors to provide teachers with suggestions for constructing exam questions.
This setup simulated a realistic scenario where a large pool of candidate distractors is generated for teachers to review and select from.
We also varied the selection batch sizes to examine their impact on the quality of the distractors.

Specifically, we experimented with selection sizes of 30, 10, and 3 distractors per round to evaluate the LLM's performance under different conditions.  
As shown in Table~\ref{tab: distractors_count_per_round}, selection sizes included 30 distractors in one round, 10 distractors across three rounds, and 3 distractors across ten rounds.  
The ``selection size'' indicates the number of distractors the LLM generated per round. 
The results show that selecting 30 distractors at once does not yield the best outcomes.  
Smaller selection sizes, such as 3 or 10 distractors per round, yield better F1 and NDCG scores, indicating higher quality.  
This suggests that requesting too many distractors at once can overwhelm the LLM and reduce selection effectiveness.  
Based on these findings, selecting three distractors at a time is the most effective strategy for the ISSR framework.

\section{Conclusion}
Vocabulary acquisition is fundamental to mastering second languages, as a rich vocabulary enhances both comprehension and expression. 
Designing effective vocabulary tests is crucial in helping learners consolidate their understanding and identify gaps in their knowledge. 
In vocabulary tests, teachers must often invest significant time and effort into generating suitable distractors---options that are misleading but cannot serve as correct answers.
However, this process is labor-intensive and prone to inconsistencies. 
Therefore, the ability to automatically generate plausible distractors is of great importance in reducing the workload for teachers and enhancing the quality of the tests. 
These automatically generated distractors help assess whether learners truly understand the meaning of the target words, rather than relying on mere guessing.

In this study, we present the ISSR framework to assist teachers in generating suitable distractors for vocabulary test design.
To achieve this goal, we analyze vocabulary questions used in actual exams and examine the relationship between the target word and distractors. 
We also investigate the factors that make a distractor appealing to students and develop a set of predefined filtering rules to enhance distractor quality.
Next, we explore the capabilities of LLMs in automatically generating distractors for vocabulary questions.
Our findings indicate that LLMs perform better when provided with a set of distractor candidates to choose from, rather than generating distractors directly, due to the instability of LLMs and limitations in producing large numbers of distractors.
To address these issues, we propose a distractor selector, a module that leverages LLMs to select plausible distractors from the generated candidates.
Finally, we introduce a distractor validator with a self-review mechanism that leverages the LLM's ability to solve vocabulary questions. 
This mechanism filters out distractors that fit the stem but could lead to multiple valid answers and thus render the question invalid.

Although we find that the proper use of LLMs enhances the ability to automatically generate vocabulary question distractors, ISSR still has some drawbacks.
Since ISSR leverages both the LLM and BERT simultaneously, it requires significantly more computing resources than similar work.
Additionally, because the self-review mechanism involves converting questions into binary choices to individually verify the validity of candidate distractors, distractor generation using ISSR is slow.
Addressing the limitations in computing resources and generation efficiency is left as future work.
Moreover, the criteria and findings derived from the GSAT dataset are tailored to its specific characteristics, limiting generalizability to other exam contexts. 
The effectiveness of distractor selection is constrained by the initial quality of the candidate set, highlighting a bottleneck in the process. 
Future work should explore improved candidate generation methods and diverse prompting strategies to enhance LLM performance and applicability.

\bibliographystyle{elsarticle-harv}
\bibliography{elsarticle-harv}

\appendix

\section{Challenges of Polysemous Words in English Learning for Asian Students}\label{sec:translation}
One intuitive way students attempt to understand English words is by using translation software to translate the words into their native language.  
As translation software typically provides the most frequent or widely used meaning, students may become familiar only with the most common interpretation of the word.  
This reliance on translation software can lead to overlooking less common meanings, making it harder for students to recognize the word in diverse contexts.  

In this study, we seek to investigate whether it is particularly challenging for students when the correct answer in a vocabulary question is a polysemous word. 
Specifically, if a question tests a less common meaning of a polysemous word, are students more likely to choose the wrong answer due to unfamiliarity with that particular meaning? 
To explore this, we experimented to assess whether the Chinese meanings of English vocabulary words (i.e., correct answers to exam questions) align with those generated by translation tools.

First, we consulted WordNet to obtain all possible definitions of the target word.
We then combined these definitions with the question stems and used \texttt{GPT-3.5-turbo-0613} to assess whether the meaning of the target word, as used in the question stem, aligned with any of its definitions.
This process enabled us to identify which specific definition of the target word was being tested in the question.
Next, we collected publicly available test explanations from textbooks\footnote{\url{https://elearning.sanmin.com.tw/englishsite/download/download.htm}} which provide Chinese translations of the target words. These translations were written by teachers or English education experts who analyze the exam content and compile it into textbooks.
Since teachers translate the target word into its appropriate Chinese meaning based on the context of the stem, we sought to assess whether the Chinese translation corresponds to the meaning of the English target word.
Thus, we translated the Chinese meaning provided in the textbooks back into English.
We then compared this back-translated word to the original target word to see whether they corresponded.
A back-translated word from Google Translate that differs significantly from the original target word may indicate that the target word is not commonly used to represent this specific Chinese meaning.
For translations that did not match the target word, we further examined whether the meaning of the back-translated word aligned with the definition of the target word in this context.
We asked \texttt{GPT-3.5-turbo-0613} to verify whether the back-translated word corresponded to the definitions initially obtained.
If the back-translated word did not encompass the tested definition, it is likely that the target word is not commonly used to express this specific meaning in Chinese, resulting in a semantic mismatch.

The premise of this experiment is that students often rely on translation software to translate texts and learn English vocabulary.
Therefore, we sought to investigate whether the word generated by the translation software differs from the answer used in the exam.
For questions where the back-translated word did not match the definition of the answer, the average pass rate was 55.91\%, whereas for questions where the back-translated word matched, the average pass rate was 60.01\%.
This suggests that students are more likely to answer incorrectly when a polysemous word is used in a less common sense, which may not be accurately captured by translation software. 
Specifically, in questions where the answer is used in an uncommon meaning, the pass rate was 4.10\% lower than the overall average.

\section{Prompts used in this work}\label{sec:prompts}

\begin{table}[t]
    \centering
    \scriptsize
    \begin{tabular}{p{13.2cm}}
        \toprule
        \textbf{Input:} **Original Sentence**\\
        Posters of the local rock band were displayed in store windows to promote the sale of their \_\_\_\_\_ tickets.\\

        **Target Word**\\
        concert\\
    
        **Candidate Pool**\\
        ``sports'', ``proper'', ``regular'', ``personal'', ``clothes'', ``favorite'', ``traffic'', ``traditional'', ``valuable'', ``available'', ``travel``, ``necessary``, ``fashionable'', ``record'', ``official'', ``final'', ``usual'', ``clothing'', ``educational'', ``fashion'', ``journey''\\
        pick three distractors from **Candidate Pool** for stem given in Original Sentence, response each distractors per line, and starts with enumerate number. \\\hline
        \textbf{Output:} 1. journey\\
        2. traffic\\
        3. record\\
        \bottomrule
    \end{tabular}
    \caption{Prompt used in distractor selector}\label{tab:prompt_distractor_selector}
\end{table}

\begin{table}[t]
    \centering
    \scriptsize
    \begin{tabular}{p{13.2cm}}
        \toprule
        \textbf{[Binary Choice Validation for Distractor Suitability]}\\
        \textbf{Input:} Imagine you are a high school student that studying english, and you are answering question given below:\\
        The following is a vocabulary test that requires selecting one answer from given options to fill in the blank.\\
        Please select the option that fit the context best from below, response with the correct option directly, if you think both options are suitable for the context, response with ``BOTH ARE GOOD''.\\
        Question:\\
        The newcomer speaks with a strong Irish \_\_\_\_\_; he must be from Ireland.\\
        options:
        identity\\
        accent\\
        \bottomrule
        \textbf{[Independent Suitability Judgment for Distractor Validation]}\\
        \textbf{Input:} Imagine you are a english teacher that designing a vocabulary test to a second language learner, and you came up with a distractor candidate ``identity''.\\
    Qustion:\\
    The newcomer speaks with a strong Irish \_\_\_\_\_; he must be from Ireland.\\
    Correct answer:\\
    accent\\\\
    Distractor candidate:\\
    identity\\\\
    The criteria for question creation are as follows:\\
    1. The length difference between the answer and the distractor
should not exceed 2 characters.\\
    2. The answer and the distractor should share the same part
of speech.\\
    3. The difficulty levels between the answer and distractor
should be closely matched\\
    Do you think whether word ``identity'' is a good distractor or not? Response with Yes or No only.\\
    \bottomrule
    \textbf{[Semantic Consistency Check for Distractor Validation]}\\
You will now see two sentences with only one word difference between them:\\
Sentence 1:\\
The newcomer speaks with a strong Irish identity; he must be from Ireland.\\
Sentence 2:\\
The newcomer speaks with a strong Irish accent; he must be from Ireland.\\
Do these two sentences have the same meaning? Please respond with `Yes' or `No' only\\
    \bottomrule
    \end{tabular}
    \caption{Prompt used in self review}\label{tab:prompt_self_review}
\end{table}

In this section, we present the prompts utilized in the ISSR framework and experiments.
The prompt used for the distractor selector is detailed in Table~\ref{tab:prompt_distractor_selector}.
Table~\ref{tab:prompt_self_review} presents the prompts employed for the self-review mechanism.
Note that we have integrated binary choice validation for determining distractor suitability within our ISSR framework, as this method effectively filters out invalid distractors with the highest accuracy.

\section{Examples of direct generation of distractors with LLMs}\label{sec:appendix_direct_inference}

\begin{table}[!h]
    \centering
    \scriptsize
    \begin{tabular}{p{13.2cm}}
        \toprule
        \textbf{Input:} **Original Sentence** $\rightarrow$ \color{gray}{[\texttt{Stem}]} \\
        The newcomer speaks with a strong Irish \_\_\_\_\_; he must be from Ireland.\\
        **Target Word** $\rightarrow$ \color{gray}{[\texttt{Target Word}]} \\
        accent\\
        generate 3 distractors for stem given in Original Sentence, restrain your output in following format given below (for example, your generated 3 distractors are: apple, banana, orange, ...)\\
        1. apple\\
        2. banana\\
        3. orange $\rightarrow$ \color{gray}{[\texttt{Request}]}\\
        \textbf{Output:}1. brogue 2. lilt 3. twang\\
        \bottomrule
\textbf{Input:} \color{gray}{[\texttt{Same Stem}]} \\
\color{gray}{[\texttt{Same Target Word}]} \\
**Words to avoid picking**\\
brogue, twang, lilt\\
\color{gray}{[\texttt{Same Request}]} \\
\textbf{Output:}1. dialect 2. intonation 3. inflection\\
        \bottomrule
\textbf{Input:} \color{gray}{[\texttt{Same Stem}]} \\
\color{gray}{[\texttt{Same Target Word}]} \\
**Words to avoid picking**\\
dialect, lilt, brogue, inflection, intonation, twang\\
\color{gray}{[\texttt{Same Request}]} \\
\textbf{Output:}1. lilt 2. twang 3. inflection\\
        \bottomrule
\textbf{Input:} \color{gray}{[\texttt{Same Stem}]} \\
\color{gray}{[\texttt{Same Target Word}]} \\
**Words to avoid picking**\\
dialect, lilt, brogue, intonation, inflection, twang\\
\color{gray}{[\texttt{Same Request}]} \\
\textbf{Output:}1. tone 2. speech 3. drawl\\
    \bottomrule
    
\textbf{Input:} \color{gray}{[\texttt{Same Stem}]} \\
\color{gray}{[\texttt{Same Target Word}]} \\
**Words to avoid picking**\\
dialect, speech, tone, lilt, brogue, drawl, inflection, intonation, twang\\
\color{gray}{[\texttt{Same Request}]} \\
\textbf{Output:}1. lilt 2. brogue 3. twang\\
    \bottomrule

\textbf{Input:} \color{gray}{[\texttt{Same Stem}]} \\
\color{gray}{[\texttt{Same Target Word}]} \\
**Words to avoid picking**\\
dialect, speech, tone, lilt, brogue, drawl, inflection, intonation, twang\\
\color{gray}{[\texttt{Same Request}]} \\
\textbf{Output:}1. lilt 2. brogue 3. twang\\
    \bottomrule
    \end{tabular}
    \caption{Directly using GPT-3.5 to generate large quantity of distractors with small generate count per round}
    \label{tab:appendix_example_small_count_per_round}
\end{table}

\begin{table}[!h]
    \centering
    \scriptsize
    \begin{tabular}{p{13.2cm}}
        \toprule
\textbf{Input:} **Original Sentence**\\ 
In the cross-lake swimming race, a boat will be \_\_\_\_\_ in case of an emergency.\\
**Target Word**\\
standingby\\
**Words to avoid picking**\\
brogue, twang, lilt\\
generate 30 distractors for stem given in Original Sentence, restrain your output in following format given below (for example, your first three generated distractors are: apple, banana, orange)\\
1. apple\\
2. banana\\
3. orange\\
\textbf{Output:}1. ready\\
2. available\\
3. prepared\\
4. on standby\\
5. waiting\\
6. on call\\
7. on alert\\
8. on hand\\
9. at the ready\\
10. on standby\\
11. on standby\\
12. on standby\\
13. on standby\\
14. on standby\\
15. on standby\\
16. on standby\\
17. on standby\\
18. on standby\\
19. on standby\\
20. on standby\\
21. on standby\\
22. on standby\\
23. on standby\\
24. on standby\\
25. on standby\\
26. on standby\\
27. on standby\\
28. on standby\\
29. on standby\\
30. on standby\\
\bottomrule
    \end{tabular}
    \caption{Directly using GPT-3.5 to generate large quantity of distractors in single round}\label{tab:appendix_example_large_count_single_round}
\end{table}

As discussed in Section~\ref{sec:five_two}, direct inference from \texttt{gpt-3.5-turbo-0125} may result in repetitive content when generating a large number of distractors.
However, it remains unclear whether this issue arises in other LLMs.
In this section, we examine the following two issues:
(1)~The tendency of \texttt{gpt-3.5-turbo-0125} to generate repetitive content: 
We investigate patterns and underlying reasons for similar or repetitive distractors when using \texttt{gpt-3.5-turbo-0125} to generate multiple-choice questions.
(2)~Whether other LLMs also exhibit a similar tendency to generate repetitive distractors.

\noindent\textbf{Tendency of GPT-3.5-turbo to Generate Repetitive Distractors.}
We explore two intuitive methods for generating 30 distractors using LLMs: 
(1)~Instructing \texttt{GPT-3.5-turbo-0125} to generate all 30 distractors in a single round, 
and (2)~Instructing \texttt{GPT-3.5-turbo-1106} to generate 3 distractors per round, with explicit instructions to avoid repeating any previously generated distractors across rounds.

Table~\ref{tab:appendix_example_small_count_per_round} and~\ref{tab:appendix_example_large_count_single_round} present the results of the first and second methods, respectively.
The issue with the first method is not only that \texttt{gpt-3.5-turbo- 0125} generates repetitive content but also that it begins producing phrases rather than single vocabulary items, which are inappropriate for actual exams. 
This issue did not occur with the second method.

As shown in Table~\ref{tab:appendix_example_small_count_per_round}, \texttt{gpt-3.5-turbo-1106} initially generates distractors successfully and avoids repeating restricted distractors.
However, after several rounds, \texttt{gpt-3.5-turbo-1106}'s output becomes unstable, and it begins repeating distractors, even though they were restricted in the prompt.

We also found that in both methods, \texttt{gpt-3.5-turbo-0125} attempts to generate terms that fit the context, which may compromise the validity of the vocabulary questions.
Although this issue could be mitigated by using different prompts or leveraging in-context learning, it remains uncertain whether the LLM can consistently avoid generating invalid distractors.
Thus, we conclude that directly generating distractors via \texttt{gpt-3.5-turbo-0125} is ineffective due to the above challenges.

\begin{table}[!h]
    \centering
    \scriptsize
    \setlength\tabcolsep{5.5pt}
    \begin{tabular}{p{13.2cm}}
        \toprule
\textbf{Input:} **Original Sentence**\\ 
In the cross-lake swimming race, a boat will be \_\_\_\_\_ in case of an emergency.\\
**Target Word**\\
standingby\\
**Words to avoid picking**\\
brogue, twang, lilt\\
generate 30 distractors for stem given in Original Sentence, restrain your output in following format given below (for example, your first three generated distractors are: apple, banana, orange)\\
1. apple\\
2. banana\\
3. orange\\
\textbf{Output:}Here are 30 distractors for the target word "accent" in the given sentence:
1. dialect \\
2. brogue \\
3. tone \\
4. inflection \\
5. pitch \\
6. timbre \\
7. voice \\
8. pronunciation \\
9. lilt \\
10. idioms \\
11. slang \\
12. drawl \\
13. twang \\
14. cadence \\
15. rhythm \\
16. emphasis \\
17. mannerism \\
18. speech \\
19. articulation \\
20. enunciation \\
21. modulation \\
22. expression \\
23. phraseology \\
24. parlance \\
25. vocabulary \\
26. elocution \\
27. diction \\
28. syntax \\
29. utterance \\
30. intonation \\

\bottomrule
    \end{tabular}
    \caption{Directly using LLAMA3.1-70B to generate large quantity of distractors in single round}\label{tab:appendix_llama3_70b_example_large_count_single_round}
\end{table}

\begin{table}[!h]
    \centering
    \scriptsize
    \begin{tabular}{p{13.2cm}}
        \toprule
        \textbf{Input:} **Original Sentence** $\rightarrow$ \color{gray}{[\texttt{Stem}]} \\
        The newcomer speaks with a strong Irish \_\_\_\_\_; he must be from Ireland.\\
        **Target Word** $\rightarrow$ \color{gray}{[\texttt{Target Word}]} \\
        accent\\
        generate 3 distractors for stem given in Original Sentence, restrain your output in following format given below (for example, your generated 3 distractors are: apple, banana, orange, ...)\\
        1. apple\\
        2. banana\\
        3. orange $\rightarrow$ \color{gray}{[\texttt{Request}]}\\
        \textbf{Output:}1. brogue 2. dialect 3.tone\\
        \bottomrule
\textbf{Input:} \color{gray}{[\texttt{Same Stem}]} \\
\color{gray}{[\texttt{Same Target Word}]} \\
**Words to avoid picking**\\
brogue, dialect, tone\\
\color{gray}{[\texttt{Same Request}]} \\
\textbf{Output:}1. inflection 2. lilt 3. twang\\
        \bottomrule
\textbf{Input:} \color{gray}{[\texttt{Same Stem}]} \\
\color{gray}{[\texttt{Same Target Word}]} \\
**Words to avoid picking**\\
brogue, dialect, tone, inflection, lilt, twang\\
\color{gray}{[\texttt{Same Request}]} \\
\textbf{Output:}1. pronunciation 2. cadence 3. drawl\\
        \bottomrule
\textbf{Input:} \color{gray}{[\texttt{Same Stem}]} \\
\color{gray}{[\texttt{Same Target Word}]} \\
**Words to avoid picking**\\
brogue, dialect, tone, inflection, lilt, twang, pronunciation, cadence, drawl\\
\color{gray}{[\texttt{Same Request}]} \\
\textbf{Output:}1. timbre 2. pitch 3. intonation\\
    \bottomrule
    \end{tabular}
    \caption{Directly using LLAMA3.1-70B to generate large quantity of distractors with small generate count per round}\label{tab:appendix_llama3_70b_example_small_count_per_round}
\end{table}

\noindent \textbf{Exploring Whether Other LLMs Exhibit a Tendency to Generate Repetitive Distractors.}
We further examined the performance of Llama3 8B and Llama3 70B in generating a large number of distractors, employing the same method used with \texttt{GPT-3.5-turbo}. 
Table~\ref{tab:appendix_llama3_70b_example_large_count_single_round} presents an example in which Llama3 70B successfully generates a large number of distractors in a single round, and Table~\ref{tab:appendix_llama3_70b_example_small_count_per_round} shows the results when generating a smaller number of distractors per round, repeated until the desired number of distractors is achieved.
These examples demonstrate that Llama3 70B successfully performs the task, producing plausible distractors that meet the requirements of the exam.

However, Llama3 8B struggles with this task.
As shown in Table~\ref{tab:appendix_llama3_8b_example_large_count_single_round}, attempting to generate a large number of distractors in a single round often results in repetitive output.
Similarly, Table~\ref{tab:appendix_llama3_8b_example_small_count_per_round} shows that even when generating a small number of distractors per round, Llama3 8B encounters similar challenges to \texttt{gpt-3.5-turbo-1106} in producing consistent, contextually relevant distractors.

Although models such as Llama3 70B demonstrate reliable performance in generating multiple distractors with both methods, the results suggest the performance is strongly influenced by the model size.
To enhance the stability and quality of distractor generation across different LLMs, we propose an alternative approach called \textit{distractor selection} in ISSR: instead of directly generating distractors, the model selects plausible distractors from a predefined candidate set, potentially improving consistency and accuracy.

\begin{table}[!h]
    \centering
    \scriptsize
    \setlength\tabcolsep{5.5pt}
    \begin{tabular}{p{13.2cm}}
        \toprule
\textbf{Input:} **Original Sentence**\\ 
In the cross-lake swimming race, a boat will be \_\_\_\_\_ in case of an emergency.\\
**Target Word**\\
standingby\\
**Words to avoid picking**\\
brogue, twang, lilt\\
generate 30 distractors for stem given in Original Sentence, restrain your output in following format given below (for example, your first three generated distractors are: apple, banana, orange)\\
1. apple\\
2. banana\\
3. orange\\
\textbf{Output:}To generate distractors for the given sentence, we'll consider words that fit grammatically and contextually. The target word is "standingby," which means ready and prepared to act in case of an emergency.

Distractors:

1. anchored \\
2. drifting \\
3. floating \\
4. moored \\
5. docked \\
6. beached \\
7. stranded \\
8. afloat \\
9. buoyed \\
10. submerged \\
11. grounded \\
12. marooned \\
13. castaway \\
14. wrecked \\
15. sunken \\
16. listing \\
17. tilting \\
18. keeling \\
19. capsized \\
20. overturned \\
21. keeled \\
22. floundering \\
23. foundering \\
24. listing \\
25. tilting \\
26. aghast \\
27. shocked \\
28. stunned \\
29. numbed \\
30. dazed \\

\bottomrule
    \end{tabular}
    \caption{Directly using LLAMA3.1-8B to generate large quantity of distractors in single round}\label{tab:appendix_llama3_8b_example_large_count_single_round}
\end{table}

\begin{table}[!h]
    \centering
    \scriptsize
    \setlength\tabcolsep{5.5pt}
    \begin{tabular}{p{13.2cm}}
        \toprule
        \textbf{Input:} **Original Sentence** $\rightarrow$ \color{gray}{[\texttt{Stem}]} \\
        The newcomer speaks with a strong Irish \_\_\_\_\_; he must be from Ireland.\\
        **Target Word** $\rightarrow$ \color{gray}{[\texttt{Target Word}]} \\
        accent\\
        generate 3 distractors for stem given in Original Sentence, restrain your output in following format given below (for example, your generated 3 distractors are: apple, banana, orange, ...)\\
        1. apple\\
        2. banana\\
        3. orange $\rightarrow$ \color{gray}{[\texttt{Request}]}\\
        \textbf{Output:}1. brogue 2. twang 3. dialect\\
        \bottomrule
\textbf{Input:} \color{gray}{[\texttt{Same Stem}]} \\
\color{gray}{[\texttt{Same Target Word}]} \\
**Words to avoid picking** \\
brogue, dialect, twang
\color{gray}{[\texttt{Same Request}]} \\
\textbf{Output:}1. intonation 2. cadence 3. modulation\\
        \bottomrule
\textbf{Input:} \color{gray}{[\texttt{Same Stem}]} \\
\color{gray}{[\texttt{Same Target Word}]} \\
**Words to avoid picking**\\
brogue, dialect, twang, intonation, cadence, modulation\\
\color{gray}{[\texttt{Same Request}]} \\
\textbf{Output:}1. brogue 2. dialect 3. cadence\\
        \bottomrule
\textbf{Input:} \color{gray}{[\texttt{Same Stem}]} \\
\color{gray}{[\texttt{Same Target Word}]} \\
**Words to avoid picking**\\
brogue, intonation, dialect, twang, cadence, modulation\\ 
\color{gray}{[\texttt{Same Request}]} \\
\textbf{Output:}1. pronunciation 2. cadence 3. drawl\\
        \bottomrule
\textbf{Input:} \color{gray}{[\texttt{Same Stem}]} \\
\color{gray}{[\texttt{Same Target Word}]} \\
**Words to avoid picking**\\
brogue, intonation, dialect, twang, cadence, modulation, pronunciation, drawl\\
\color{gray}{[\texttt{Same Request}]} \\
\textbf{Output:}1. drawl 2. brogue 3. cadence\\

\bottomrule
    \end{tabular}
    \caption{Directly using LLAMA3.1-8B to generate large quantity of distractors with small generate count per round}\label{tab:appendix_llama3_8b_example_small_count_per_round}
\end{table}

\section{Human Evaluation of ISSR-Generated Questions}
\label{sec:human_eval}
In our experiments, we demonstrated that ISSR outperforms other baseline models.  
However, it remains uncertain whether the distractors generated by ISSR are both plausible enough and valid enough to effectively challenge examinees.  
To evaluate this, we conducted a study using 30 questions from the GSAT exams, where ISSR-generated distractors were used.  
Thirteen university students with sufficient English proficiency for the GSAT test and strong prior GSAT English performance were invited to participate.  
The questions were selected based on their original pass rates.  
We first sorted the questions by pass rates, divided them into three equal-sized groups, and randomly selected 10 questions from each group.  
The groups were organized as follows:  
\begin{itemize}
  \item \textbf{First Group:} This group included the questions with the lowest pass rates, ranging from a minimum pass rate of 23\% to a maximum of 54\%.
  The standard deviation of pass rates within this group are 7.15.
  \item \textbf{Second Group:} The questions in this group had pass rates ranging from 54\% to 65\%, with a standard deviation of 3.27.
  \item \textbf{Third Group:} This group covered questions with the highest pass rates, from a minimum of 65\% to a maximum of 87\%.
The standard deviation for this group was 5.30.
\end{itemize}

This approach allowed us to assess the performance of ISSR-generated distractors across questions of varying difficulty levels, providing insight into how effectively ISSR-generated distractors function in both easier and more challenging contexts.
Additionally, we asked the participants to label any distractors they confidently identified as also plausible as correct answers, aside from the most accurate option, which would render the question invalid.
We manually verified whether these labeled distractors could indeed serve as correct answers.
Our analysis showed that out of 90 generated distractors, 8 were labeled as correct answers, potentially rendering the questions invalid.
In one case, a student labeled the correct answer as an alternative plausible option while selecting a distractor as the intended answer.
This indicates that ISSR-generated distractors generally support question design by ensuring only one correct answer per question.

Table~\ref{tab: human_eva} presents the students' accuracy.  
The ``Challenging questions'' column indicates the number of questions labeled as confusing due to the generated distractors. 
The results show that over half of the students achieved accuracy between 90 and 100.  
Among these, 6 questions were identified as ``plausible questions,'' where the distractors effectively misled students.  
Despite their strong English vocabulary background, these students still found 6 questions challenging, demonstrating ISSR's effectiveness in generating realistic and confusing distractors.  
Two examinees with accuracies between 60 and 70 did not label any distractors as confusing, likely due to difficulty in recognizing vocabulary, which may have limited their ability to assess distractor effectiveness.  
Based on these observations, we conclude that ISSR-generated distractors are effective at subtly challenging students, particularly those with moderate to high proficiency.

\begin{table}[t]
\centering
\small
\begin{tabular}{lrr}\toprule
Accuracy (\%) & Students & Challenging questions\\\midrule
90--100 &7 & 6 \\
80--90 &3&4 \\
70--80 &0&0 \\
60--70 &2&0 \\
\bottomrule
\end{tabular}
\caption{Student accuracy on ISSR generated distractors}\label{tab: human_eva}
\end{table}

\section{LLM Evaluation on English Vocabulary Test}
\label{sec:LLM_answer_ability}

\begin{table}[t]
\centering
\small
\begin{tabular}{lrr}\toprule
LLM &Accuracy \\\midrule
Llama3-70B &98.46\% \\
Llama3-8B &95.90\% \\
GPT-3.5 &95.90\% \\
GPT-4 &\textbf{100}\% \\
\bottomrule
\end{tabular}
\caption{Performance of various LLMs in answering vocabulary questions}\label{tab: temp5}
\end{table}

We evaluate the LLM's performance on a vocabulary test because its ability to correctly answer these questions is fundamental to the distractor generation method. 
The LLM must be able to accurately distinguish between the correct answer and the distractors to ensure that the generated distractors are not mistakenly identified as valid answers, thus maintaining the requirement for a single correct answer in the question design.
If the LLM incorrectly selects a distractor as the answer, it is highly likely that the distractor also fits the stem, thereby invalidating the generated vocabulary question.
To assess this further, we conducted an experiment using the GSAT English test, where LLMs were presented with actual test questions and tasked with selecting the most appropriate answer from a set of well-designed plausible distractors.
Table~\ref{tab: temp5} shows the accuracy of various LLMs in answering English vocabulary questions.
The results indicate that LLMs possess a strong capability in solving well-designed vocabulary problems.
To sum up, we propose that the self-review mechanism may be effective, as the LLM can accurately select the correct answer without mistakenly choosing plausible distractors.

\end{document}